\def\year{2021}\relax
\documentclass[letterpaper]{article} 
\usepackage{aaai21}  
\usepackage{times}  
\usepackage{helvet} 
\usepackage{courier}  
\usepackage[hyphens]{url}  
\usepackage{graphicx} 
\usepackage{natbib}  
\usepackage{caption} 
\usepackage{tabulary}
\usepackage{booktabs}

\usepackage{eqnarray,amsmath}
\usepackage{bm}
\usepackage{amssymb}
\usepackage{multirow}
\usepackage{tabulary}
\usepackage{makecell}

\usepackage{graphicx}
\usepackage[switch]{lineno}
\title{Scheduled Sampling in Vision-Language Pretraining\\ with Decoupled Encoder-Decoder Network}
\author{Yehao Li \textsuperscript{\rm 1}, Yingwei Pan \textsuperscript{\rm 1}, Ting Yao \textsuperscript{\rm 1}, Jingwen Chen \textsuperscript{\rm 2}, Tao Mei \textsuperscript{\rm 1}\\}
\affiliations{\textsuperscript{\rm 1} JD AI Research, Beijing, China\\
\textsuperscript{\rm 2} Sun Yat-sen University, Guangzhou, China\\
\{yehaoli.sysu, panyw.ustc, tingyao.ustc, chenjingwen.sysu\}@gmail.com, tmei@jd.com}

\begin{document}

\maketitle
\begin{abstract}
  Despite having impressive vision-language (VL) pretraining with BERT-based encoder for VL understanding, the pretraining of a universal encoder-decoder for both VL understanding and generation remains challenging. The difficulty originates from the inherently different peculiarities of the two disciplines, e.g., VL understanding tasks capitalize on the unrestricted message passing across modalities, while generation tasks only employ visual-to-textual message passing. In this paper, we start with a two-stream decoupled design of encoder-decoder structure, in which two decoupled cross-modal encoder and decoder are involved to separately perform each type of proxy tasks, for simultaneous VL understanding and generation pretraining. Moreover, for VL pretraining, the dominant way is to replace some input visual/word tokens with mask tokens and enforce the multi-modal encoder/decoder to reconstruct the original tokens, but no mask token is involved when fine-tuning on downstream tasks. As an alternative, we propose a primary scheduled sampling strategy that elegantly mitigates such discrepancy via pretraining encoder-decoder in a two-pass manner. Extensive experiments demonstrate the compelling generalizability of our pretrained encoder-decoder by fine-tuning on four VL understanding and generation downstream tasks. Source code is available at \url{https://github.com/YehLi/TDEN}.
\end{abstract}

\section{Introduction}

Vision and language are two fundamental capabilities of human intelligence. The interactions in between support the series of uniquely human capacity, such as visual-language (VL) understanding (e.g., visual question answering \cite{antol2015vqa,anderson2017bottom}), and VL generation (e.g., image captioning \cite{vinyals2015show,pan2020x,li2019pointing,yao2017incorporating} and video captioning \cite{chen2019temporal,li2018jointly,xu2016msr,pan2016jointly,pan2017video}). Inspired by recent development of natural language pretraining (e.g., BERT \cite{devlin2019bert}), there has been a steady momentum of breakthroughs that push the limits of VL tasks via VL pretraining. Most of existing VL pretraining techniques (e.g., ViLBERT \cite{lu2019vilbert}, LXMERT \cite{tan2019lxmert}, VL-BERT \cite{su2019vl}) focus on learning a universal multi-modal encoder for VL understanding downstream tasks, which can not be directly applied to VL generation task as the lack of a decoder for language modeling.
Nevertheless, how to learn a pre-trainable encoder-decoder over large-scale VL benchmarks that can be fine-tuned on both VL understanding and generation downstream tasks is a seldom explored territory.

In this work, we target for pretraining a universal encoder-decoder structure, that facilitates both VL understanding and generation downstream tasks. This direction has recently been studied in Unified VLP \cite{zhou2019unified} by using a single-stream BERT-based encoder-decoder with a joint pretraining of VL understanding proxy task (MLM: masked language modeling) and generation proxy task (MSG: masked sentence generation). Nevertheless, the design of single-stream structure processes both modality inputs via the same transformer blocks, leaving the inherent different peculiarity of each modality and each VL proxy task not fully exploited. That severely limits the generalization of pre-trained encoder-decoder across different kinds of VL downstream tasks (as observed in \cite{zhou2019unified} that joint pre-training with MLM \& MSG is less effective than separate pre-training with MLM/MSG~on each downstream task). Instead, this paper considers a two-stream decoupled design of encoder-decoder, aiming to exploit the mutual relationship between different modalities/VL proxy tasks for enhancing the robustness on VL downstream tasks. Our design, the Two-stream Decoupled Encoder-decoder Network (\textbf{TDEN}), consists of two encoders to process each modality inputs, together with the decoupled cross-modal encoder and decoder for each kind of proxy task. In particular, the shared object/sentence encoder independently learns the representations of each modality via intra-interaction, which offers a fertile ground for multi-modal reasoning that supports both VL understanding and generation. Moreover,~during pretraining, MLM makes the predictions of masked word/visual tokens in a single shot based on unrestricted message passing between two modalities, while MSG requires the auto-regressive reconstruction of input sentence and solely triggers the visual-to-textual message passing. In view of such fundamental differences in between, we involve two decoupled cross-modal encoder and decoder that separately perform each kind of proxy task in a multi-task fashion.

\begin{figure*}[!tb]
\centering {\includegraphics[width=0.82\textwidth]{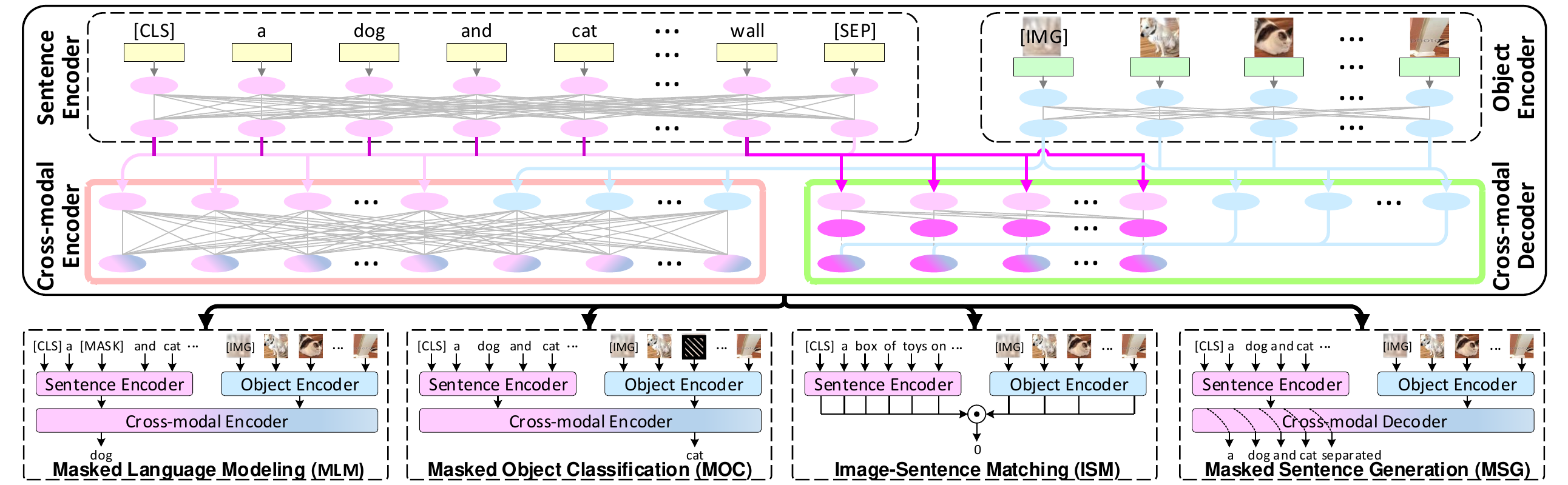}}
\caption{\small \textbf{Top}: An overview of our TDEN with a two-stream decoupled architecture design, consisting of four modules (i.e., the shared object and sentence encoders for contextually encoding each modality inputs, cross-modal encoder for VL understanding, and cross-modal~decoder for VL generation. \textbf{Below}: Four VL proxy tasks in TDEN, i.e., MLM, MOC, and ISM for VL understanding, and MSG for VL generation.}
\label{fig:figframework}
\end{figure*}

In the meantime, we notice that several VL proxy tasks, e.g., MLM and masked object classification (MOC), have become a default choice for VL pretraining, which replace some input word/visual tokens with mask tokens (\texttt{[MASK]}) and enforce the network to recover the primary inputs. Though promising results are reported on downstream tasks, these mask-based VL proxy tasks inevitably introduce a discrepancy
at network fine-tuning for downstream tasks, where no artificial mask token is involved. To mitigate such discrepancy (in analogy to exposure bias \cite{bengio2015scheduled} in RNN based sequence-to-sequence generation), we derive a particular form of scheduled sampling for VL pretraining. Technically, a two-pass pretraining scheme is devised to enable scheduled sampling in our BERT-based encoder-decoder structure. Our launching point is to additionally formulate a more practical pretraining pass by replacing the artificial mask tokens with the generated real ones, which are directly sampled from the previous predictions of cross-modal encoder/decoder in the prior pass.

The main contribution of this work is the proposal of a universal encoder-decoder structure that facilitates both VL understanding and generation tasks. This also leads to the elegant view of how a pre-trainable encoder-decoder structure should be designed for fully exploiting the mutual but also fuzzy relations between different modalities and VL proxy tasks, and how to bridge the discrepancy between pretraining and finetuning tailored to VL pretraining. Through an~extensive set of experiments on four VL understanding and generation downstream tasks, we demonstrate that our pre-trained TDEN achieves new state-of-the-art performances for each task.

\section{Related Work}
\noindent\textbf{Language Pretraining.}
Language pretraining has attracted increasing attention in NLP field, and obtained impressive performances in various natural language understanding tasks.
GPT \cite{radford2018improving} is one of the early successes for language pretraining by exploiting the unidirectional word context to learn general language representations. ELMo \cite{peters2018deep} also provides the context sensitive features to downstream tasks.
By integrating proxy tasks (masked language modeling and next sentence prediction) for pretraining, BERT \cite{devlin2019bert} enables the learning of bidirectional representations. Recently, XLNet \cite{yang2019xlnet} upgrades BERT with a generalized autoregressive pretraining mechanism. Unlike BERT-type pretraining models which only support language understanding via one encoder, several recent works go beyond the traditional language pretraining and aim to pretrain an encoder-decoder network for language generation through generative proxy tasks (e.g., masked sequence to sequence learning in MASS \cite{song2019mass}, sequence-to-sequence modeling in UNILM \cite{dong2019unified}, and denoising sequence-to-sequence modeling in BART \cite{lewis2019bart}). Our work pursuits their vision-language counterpart by pretraining a universal encoder-decoder structure and finetuning it on both VL understanding and generation tasks.

\begin{figure*}[!tb]
\centering {\includegraphics[width=0.86\textwidth]{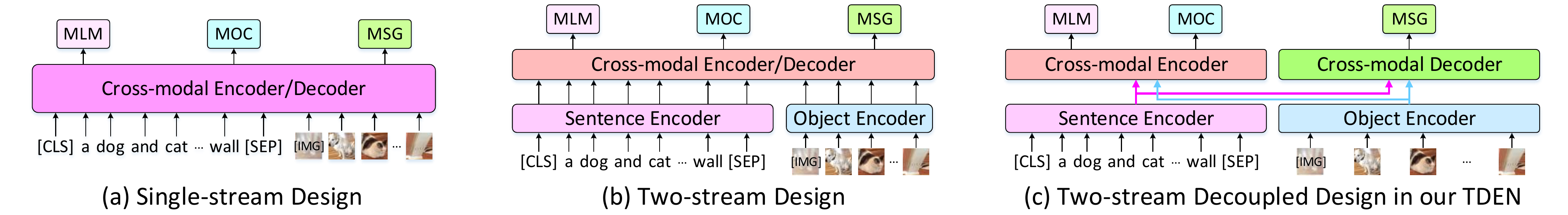}}
\caption{\small Three architectures for VL pretraining: (a) single-stream design (e.g., VL-BERT \cite{su2019vl}, Unified VLP \cite{zhou2019unified}); (b) two-stream design (e.g., ViLBERT \cite{lu2019vilbert}, LXMERT \cite{tan2019lxmert}); (c) two-steam decoupled design (our~TDEN).}
\label{fig:figarchitecture}
\end{figure*}

\noindent\textbf{Vision-Language Pretraining.}
Inspired by language pretraining, the research community starts to pay more attention to VL pretraining in multi-modal scenario.
In particular, VisualBERT \cite{li2019visualbert} directly migrates BERT to VL pretraining by involving visually-grounded proxy tasks (e.g., MLM coupled with image and image-sentence matching (ISM)). After that, a new VL understanding proxy task, named masked object classification (MOC), is widely adopted in VL pretraining techniques (UNITER \cite{chen2019uniter}, Unicoder-VL \cite{li2019unicoder}, and VL-BERT \cite{su2019vl}), which further enhances the region-level vision-language alignment. Different from BERT-type techniques that capitalize on a single-stream encoder, ViLBERT \cite{lu2019vilbert} and LXMERT \cite{tan2019lxmert} utilize a more detailed two-stream encoder structure for VL pretraining, which consists of two separate encoders and one cross-modal encoder. Most recently, unlike the aforementioned methods that only support VL understanding tasks via a multi-modal encoder, Unified VLP \cite{zhou2019unified} learns a single-stream encoder-decoder that can be generalized to both VL understanding and generation tasks. In this paper, we also focus on the latter challenging task in VL pretraining. Instead of learning a single-stream encoder-decoder structure in Unified VLP, we utilize a two-stream decoupled design of encoder-decoder that reflects the mutual relationship between different modalities/VL proxy tasks. In addition, a novel scheduled sampling strategy is proposed to mitigate the pretraining-finetuning discrepancy when pretraining the universal encoder-decoder.

\section{Approach}

We present a pre-trainable Two-stream Decoupled Encoder-decoder Network (TDEN) that facilitates both vision-language understanding and generation tasks, as shown in Figure \ref{fig:figframework}. In this section, we firstly elaborate the rationale behind our two-stream decoupled design of encoder-decoder, followed by a brief introduction of four VL understanding and generation proxy tasks adopted in our TDEN. Finally, a primary scheduled sampling strategy tailored to VL pretraining is described in detail.

\begin{figure*}[!tb]
\centering {\includegraphics[width=0.98\textwidth]{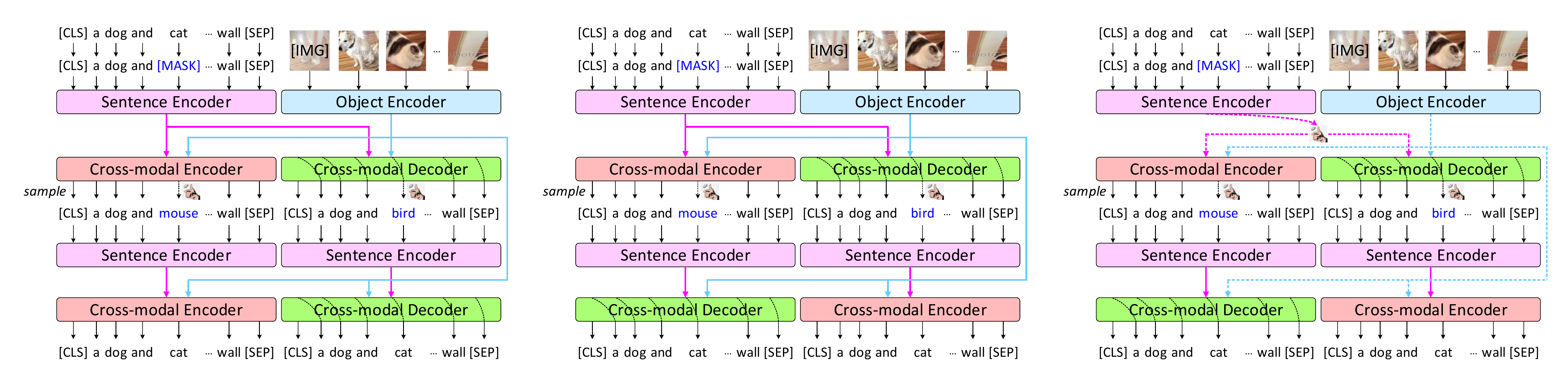}}
\caption{\small Three designs of two-pass pretraining schemes for VL pretraining with scheduled sampling. \textbf{Left (Two-Pass-A)}: In the second pass, we replace the input masked word tokens of cross-modal encoder/decoder with the generated ones sampled from the corresponding output word distributions of cross-modal encoder/decoder in the first pass. \textbf{Middle (Two-Pass-B)}: In the second pass, we exchange the input word token sequences of cross-modal encoder and decoder in Two-Pass-A. \textbf{Right (Two-Pass-C)}: Two-Pass-C is a cost-efficient version of Two-Pass-B by involving a switch between sentence encoder and cross-modal encoder/decoder in the first pass, aiming to randomly choose only the cross-modal encoder or decoder (50\% probability) for pretraining. Therefore, in the second pass, only the cross-modal decoder/encoder is activated for a more practical pretraining.}
\label{fig:training}
\end{figure*}

\subsection{Architecture Design}
\label{sec:model}

The core architectures in existing VL pretraining techniques for VL understanding tasks can be briefly grouped into two directions: single-stream design (e.g., Unicoder-VL \cite{li2019unicoder} and VL-BERT~\cite{su2019vl}) and two-stream design (e.g., ViLBERT \cite{lu2019vilbert} and LXMERT \cite{tan2019lxmert}). As depicted in Figure \ref{fig:figarchitecture} (a), the~single-stream design directly fuses the primary multi-modal inputs from the start, and leaves the inherent peculiarity of each modality unexploited. Instead, the two-stream design (Figure \ref{fig:figarchitecture} (b)) independently processes each modality via separate encoders, followed with a shared cross-modal encoder for cross-modal interaction. Such design enables a flexible and specialized encoding process for each modality, and fully modulates features of each modality into high-level semantic features~for multi-modal reasoning. Nevertheless, when directly migrating existing single-stream/two-stream design into the pretraining of encoder-decoder for both VL understanding and generation tasks (e.g., Unified VLP \cite{zhou2019unified}), two different kinds of VL proxy tasks (MLM and MSG) are enforced to be learnt through a shared cross-modal encoder-decoder. This way apparently ignores the inherently different peculiarity of each VL proxy task (e.g., MLM capitalizes on the unrestricted message passing between two modalities, while MSG only exploits the visual-to-textual message passing), and thus severely limits the generalization of pre-trained encoder-decoder across different types of VL downstream tasks.

Accordingly, we design a specific two-stream decoupled encoder-decoder structure (as shown in Figure \ref{fig:figarchitecture} (c)) with two principles: (\emph{i}) we want the encoders of each modality to be independent to each other for capturing intra-interactions, but shared across different proxy tasks to offer a fertile ground for multi-modal reasoning; (\emph{ii}) we want to decouple the two kinds of proxy tasks during pretraining by involving two decoupled cross-modal encoder and decoder, and each one is in charge of one kind of proxy task. The detailed architecture of our TDEN is illustrated as following:

\textbf{Notations.}
The inputs of our TDEN are image-sentence pairs $\{\mathcal{I}, \mathcal{S}\}$ derived from a large-scale image-sentence benchmark (Conceptual Captions \cite{sharma2018conceptual}). Each input image $\mathcal{I}$ is initially represented as a group of detected image regions (i.e., visual tokens) $\mathcal{R}_I = \left\{ \bm{r}_{i} \right\}^{N_I}_{i=1}$, which are generated via object detector (Faster R-CNN \cite{ren2015faster}). For each corresponding sentence $\mathcal{S}$, we represent it as a sequence of word tokens $\mathcal{W}_{S} = \left\{ \bm{w}_{j} \right\}^{N_S}_{j=1}$. Note that both of the input visual and word tokens are position-aware features by additionally exploiting the 2D/1D position information of each region/word as in \cite{lu2019vilbert}.

\textbf{Object Encoder and Sentence Encoder.} In our TDEN, both of object encoder and sentence encoder are implemented as a series of transformer blocks, that independently encode the inputs of each modality with intra-modal contextual information. Formally, for object encoder, we firstly integrate the primary visual tokens with a special visual token \texttt{[IMG]} (the mean-pooling object representation over all detected regions) as the input visual token sequence ($\tilde{\mathcal{R}}_I=\left\{ \bm{r}_{i} \right\}^{N_I}_{i=0}$). Here \texttt{[IMG]} indicates the beginning of visual token sequence. After that, $K_I$ stacked transformer blocks are utilized to perform self-attention over input visual token sequence, aiming to enhance each visual token representation with intra-modal context information mined from the pervious transformer block. Similarly, in sentence encoder, we leverage $K_S$ stacked transformer blocks to capture the intra-modal context information within input word token sequence. Two special tokens \texttt{[CLS]} and \texttt{[SEP]} are included to indicate the beginning and ending of the input word token sentence ($\tilde{\mathcal{W}}_{S} = \left\{ \bm{w}_{j} \right\}^{N_S+1}_{j=0}$).

\textbf{Cross-modal Encoder.} The cross-modal encoder is devised to fully capture the inter-modal interaction across different modalities for VL understanding. Technically, given the enhanced visual and word tokens from each encoder ($\mathcal{H}^0_S=\left\{ \bm{h}_{j} \right\}^{N_S+1}_{j=0}$, $\mathcal{H}^0_I=\left\{ \bm{h}_{i} \right\}^{N_I}_{i=0}$), we first concatenate them as multi-modal input: ${\mathcal{H}}^{0}_{SI} = [\mathcal{H}^{0}_S, {\mathcal{H}}^{0}_I]$. We further feed ${\mathcal{H}}^{0}_{SI}$ into a stack of $K_E$ transformer blocks, leading to the contextual multi-modal features of each visual/word token (${\mathcal{H}}^{E}_{SI} = \left\{ {\bm{h}}^{E}_{i}\right\}^{N_I+N_S+2}_{i=0}$).

\textbf{Cross-modal Decoder.} We additionally involve a cross-modal decoder that learns to auto-regressively reconstruct the input sentence word-by-word conditioned on the input image, which mimics the sequence generation process and thus supports VL generation downstream task. Here we implement the cross-modal decoder by stacking $K_D$ transformer-based decoder layers \cite{vaswani2017attention}. In particular, each transformer-based decoder layer first collects contextual information from all the ``past'' word tokens via self-attention, and then performs cross-attention over all visual tokens for next word prediction.

\subsection{VL Understanding and Generation Proxy Tasks}
\label{sec:task}
When pretraining our TDEN, we consider four specific objectives with regard to VL understanding and generation proxy tasks: masked language modeling (MLM), masked object classification (MOC), and image-sentence matching (ISM) that pursue VL alignment at both region-level and image-level for VL understanding, and masked sentence generation (MSG) that simulates the auto-regressive process of language modeling for VL generation.

\textbf{Masked Language Modeling (MLM).}
Motivated by single-modal MLM proxy task in BERT, we leverage a multi-modal version of MLM for pretraining. The objective of MLM is to recover the masked word tokens depending on the unmasked word tokens and visual tokens. In particular, we randomly mask the input word tokens (15\% probability) by replacing each masked word token with a mask token \texttt{[MASK]}. The output contextual multi-modal features of masked word tokens in cross-modal encoder are thus utilized to re-produce the original word tokens via a classifier over the whole vocabulary, driven by softmax cross-entropy~loss.

\textbf{Masked Object Classification (MOC).}
In analogy to MLM, MOC is additionally involved to enable the reconstruction of objects reflected in masked visual tokens conditioned on the unmasked visual tokens and word tokens, which further enhances the region-level VL grounding. Similarly, the input visual tokens are randomly replaced (15\% probability) with mask token. We further feed the output contextual multi-modal features of masked visual tokens into a classifier for object classification. The objective of MOC is expressed as KL divergency loss that measures the mismatch between the estimated object distribution and the ground-truth object distribution (directly obtained from the off-the-shelf object detector) for each masked visual token.

\textbf{Image-Sentence Matching (ISM).}
Different from MLM and MOC that only exploit the local multi-modal contextual information, we involve ISM to strengthen image-sentence alignment for VL understanding.
Nevertheless, the typical implementation of ISM (e.g., \cite{lu2019vilbert}), i.e., capitalizing on the outputs of cross-modal encoder to measure image-sentence similarity, would hurt performances on VL understanding tasks. As pointed in \cite{su2019vl}, one probable reason for the performance drop is that the introduced mismatched image-sentence pairs through shared cross-modal encoder would hamper the pretraining of other proxy tasks. Instead, we directly measure the image-sentence similarity for ISM based on the outputs of object and sentence encoders, which triggers an earlier image-sentence alignment and thus avoids the negative impacts of mismatched image-sentence pairs on cross-modal encoder. In particular, we feed the outputs of object and sentence encoders into an attention-based two-layer MLP \cite{zhouyucvpr2019} to calculate the image-sentence similarity. The objective of ISM is measured as triplet ranking loss, where the negative pair is generated via Multibatch strategy~\cite{tadmor2016learning}.

\textbf{Masked Sentence Generation (MSG).}\label{MSG}
Inspired by \cite{pan2020auto,zhou2019unified}, MSG is introduced to teach cross-modal decoder how to auto-regressively decode the input sentence word-by-word conditioned on input image, endowing TDEN with the capability of sentence generation. The objective of MSG is measured as the joint negative log probability for reconstructing the sequential word tokens depending on all the ``past'' word tokens and input image.

\subsection{Scheduled Sampling in VL Pretraining}

The most typical way to pretrain TDEN is to optimize the whole encoder-decoder structure with all the four objectives of VL understanding and generation proxy tasks simultaneously. Nevertheless, when taking a close look on the objectives, we find that most of them correspond to mask-driven~proxy tasks (e.g., MLM and MOC), which corrupt the multi-modal inputs by replacing the original word/visual tokens with artificial mask tokens (\texttt{[MASK]}). This inevitably results in a discrepancy at the subsequent finetuning process for downstream tasks, since no artificial mask token is involved during finetuning. Inspired by scheduled sampling in RNN based sequence modeling \cite{bengio2015scheduled}, we derive a particular form of scheduled sampling to elegantly mitigate such pretraining-finetuning discrepancy via pretraining our TDEN in a two-pass manner. Specifically, a two-pass pretraining scheme is designed to trigger scheduled sampling during pretraining. In the first pass, a standard VL pretraining is performed with the masked multi-modal inputs, leading to the estimated word distribution of each masked word token in cross-modal encoder/decoder. Next, the second pass conducts a more practical VL pretraining by replacing the masked word tokens with plausible alternatives sampled from the first pass.

The detailed implementation of our two-pass pretraining scheme can be varied by adopting different sampling sources (i.e., each alternative input word token of encoder/decoder in the second pass can be sampled from the output distribution of cross-modal encoder or decoder in the first pass). Hence here we derive three different two-pass pretraining schemes as depicted in Figure \ref{fig:training}, respectively, named as Two-Pass-A to Two-Pass-C. Detailed comparisons in between are provided as follows:

\textbf{Two-Pass-A.} In the first scheme, we perform a standard pretraining on TDEN in the first pass, and obtain the output word distribution for each position of masked word token via cross-modal encoder and decoder. After that, we corrupt the input masked word token sequence by replacing the artificial mask tokens with the generated ones sampled from the output distributions from cross-modal encoder or decoder in the first pass, leading to two unmasked word token sequences ($\mathcal{{S}}_E$, $\mathcal{{S}}_D$). Next, in the second pass, after encoding $\mathcal{{S}}_E$ via sentence encoder, we feed the outputs (coupled with the encoded visual tokens in the first pass) into cross-modal encoder to perform MLM and MOC proxy tasks. Meanwhile, the encoded word tokens of $\mathcal{{S}}_D$ are fed into cross-modal decoder to perform MSG proxy task conditioned on the input image. We jointly optimize our TDEN through the two passes and the overall objective is measured~as:
\begin{equation}
\begin{aligned}
\mathcal{L}_{TDEN}^{A} =& \mathcal{L}_{TDEN}+ \mathcal{L}_{\text{MLM}}(\mathcal{{S}}_E, \mathcal{I}) + \mathcal{L}_{\text{MOC}}(\mathcal{{S}}_E, \mathcal{I}) \\
&+ \mathcal{L}_{\text{MSG}}(\mathcal{{S}}_D, \mathcal{I}),
\end{aligned}
\end{equation}
where $\mathcal{L}_{TDEN}$ denotes the objective of a standard VL pretraining with all the four proxy tasks in the first pass, $\mathcal{L}_{\text{MLM}}$, $\mathcal{L}_{\text{MOC}}$, and $\mathcal{L}_{\text{MSG}}$ represents the objective in MLM, MOC, and MSG, respectively.

\textbf{Two-Pass-B.} Two-Pass-B is a variant of Two-Pass-A by exchanging the input unmasked word token sequences of cross-modal encoder and decoder in the second pass. That~is, we feed encoded word tokens of $\mathcal{{S}}_E$/$\mathcal{{S}}_D$ into cross-modal decoder/encoder. The overall objective is thus calculated as:
\begin{equation}
\begin{aligned}
\mathcal{L}_{TDEN}^{B} =& \mathcal{L}_{TDEN} + \mathcal{L}_{\text{MSG}}(\mathcal{{S}}_E, \mathcal{I}) + \mathcal{L}_{\text{MLM}}(\mathcal{{S}}_D, \mathcal{I})\\
& + \mathcal{L}_{\text{MOC}}(\mathcal{{S}}_D, \mathcal{I}).
\end{aligned}
\end{equation}

\textbf{Two-Pass-C.} Due to an additional pass without mask tokens is involved in our two-pass pretraining scheme, the pretraining cost of Two-Pass-A and Two-Pass-B is doubled against a single standard VL pretraining. Therefore, we design a cost-efficient version of our two-pass pretraining scheme (Two-Pass-C), which slightly modifies Two-Pass-B by involving a switch between sentence encoder and cross-modal encoder/decoder in the first pass. The switch is utilized to randomly choose only one path (cross-modal encoder/decoder) for pretraining in the first pass. Next, when one path (e.g., cross-modal encoder) is selected, only the corresponding path (e.g., cross-modal decoder) is activated with the input unmasked word token sequence in the second pass. As such, we effectively reduce the pretraining cost by approximately a factor of 2. Accordingly, the overall objective can be expressed~as:
\begin{equation}\scriptsize
\begin{aligned}
\mathcal{L}_{TDEN}^{C} =&\alpha [\mathcal{L}_{\text{MLM}}(\mathcal{S}, \mathcal{I}) + \mathcal{L}_{\text{MOC}}(\mathcal{S}, \mathcal{I}) + \mathcal{L}_{\text{MSG}}(\mathcal{{S}}_E, \mathcal{I})] \\
&+ (1-\alpha)[ \mathcal{L}_{\text{MSG}}(\mathcal{S}, \mathcal{I}) +\mathcal{L}_{\text{MLM}}(\mathcal{{S}}_D, \mathcal{I}) +  \mathcal{L}_{\text{MOC}}(\mathcal{{S}}_D, \mathcal{I})]\\
&+ \mathcal{L}_{\text{ISM}}(\mathcal{S}, \mathcal{I}),
\end{aligned}
\end{equation}
where $\alpha \in \{0, 1\}$ indicates the switching between cross-modal encoder and decoder in the first pass.

\section{Experiments}

\subsection{Experimental Settings}

\textbf{Pretraining Data and Details.} We conduct the experiments for pretraining over the large-scale image captioning benchmark---Conceptual Captions \cite{sharma2018conceptual}. Conceptual Captions contains 3.3 million image-sentence pairs, which are automatically collected from billions of webpages. During pretraining, the adopted off-the-shelf Faster-RCNN is pre-trained on ImageNet \cite{ImageNet} and Visual Genome \cite{krishna2017visual} as in \cite{anderson2017bottom}. At most 100 image regions with detection confidences higher than 0.2 are selected as inputs. Each input image region representation is a 2,048-dimensional vector. The number of stacked transformer-based layers in object encoder, sentence encoder, cross-modal encoder, and cross-modal decoder is set as in ViLBERT \cite{lu2019vilbert}. We implement the whole architecture with PyTorch \cite{paszke2019pytorch}, optimized with Adam \cite{kingma2014adam} on 16 Tesla P40 GPUs. The mini-batch size is 1,024 and learning rate is set as 0.0001. The maximum iteration is 10 epoches.

\begin{table*}[!tb]\scriptsize
\centering
\setlength{\extrarowheight}{0.0pt}
\setlength\tabcolsep{1.6pt}
\caption{\small Performance comparison with state-of-the-art (\textbf{SOTA}) task-specific models and VL pretraining techniques (\textbf{Pre-E}: pre-trainable encoder; \textbf{Pre-ED}: pre-trainable encoder-decoder) on four VL downstream tasks. TDEN$^-$ is one variant of our TDEN which is directly trained with task-specific training data, without pretraining.$^\star$ indicates using a stronger backbone (ResNeXt).}
\label{table:exp}
\begin{tabular}{l|l|cccc|cccc|ccc|ccc}
   \Xhline{2\arrayrulewidth}
   & \multirow{2}{*}{Model} & \multicolumn{4}{c|}{Image Captioning (test 5k/test server c40)}  & \multicolumn{4}{c|}{VQA (test-dev/test-std)} & \multicolumn{3}{c|}{CIR}  & \multicolumn{3}{c}{VCR (dev/test)} \\
   & ~ &  B@4  &  M  &  C  &  S  &  Overall  &  Yes/No  &  Number  &  Other  &  R1  &  R5  &  R10  &  Q$\rightarrow$A  &  QA$\rightarrow$R  &  Q$\rightarrow$AR  \\ \hline\hline

\multirow{8}{*}{\textbf{SOTA}}
        & BUTD \cite{anderson2017bottom}     & 36.3 & 27.7 & 120.1 & 21.4 & 65.3/65.7 & 81.8/82.2 & 44.2/43.9 & 56.1/56.3 & -    & -    & -    & -   & - & -    \\
        & AoANet \cite{huang2019attention}      & 38.9 & 29.2 & 129.8 & 22.4 & -    & -   & -   & -   & -    & -    & -    & -   & -   & -    \\
        & BAN \cite{kim2018bilinear}         & -    & -    & -     & -    & 70.0/70.4 & 85.4/85.8 & 54.0/53.7 & 60.5/60.7 & -    & -    & -    & -   & - & -    \\
        & DFAF \cite{gao2019dynamic}        & -    & -    & -     & -    & 70.2/70.3 & 86.1/-~~~~~~ & 53.3/-~~~~~~ & 60.5/-~~~~~~ & -    & -    & -    & - & -   & -    \\
        & MCAN \cite{zhouyucvpr2019}        & -    & -    & -     & -    & 70.6/70.9 & 86.8/-~~~~~~ & 53.3/-~~~~~~ & 60.7/-~~~~~~ & -    & -    & -    & - & -   & -    \\
        & SCAN \cite{lee2018stacked}        & -    & -    & -     & -    & -   & -   & -   & -   & 48.6 & 77.7 & 85.2 & -   & -   & -    \\
        & R2C \cite{zellers2019recognition} & -    & -    & -     & -    & -   & -   & -   & -   & -    & -    & -    & 63.8/65.1 & 67.2/67.3 & 43.1/44.0  \\ \hline
		& TDEN$^-$ & 38.8 & 28.8 & 128.2 & 22.8 & 69.8/-~~~~~~ & 86.3/-~~~~~~ & 49.5/-~~~~~~ & 60.4/-~~~~~~ & 55.6 & 81.8 & 88.5 & 73.0/-~~~~~~ & 74.4/-~~~~~~ & 54.5/-~~~~~~  \\ \hline
\multirow{4}{*}{\textbf{Pre-E}}
        & LXMERT \cite{tan2019lxmert}       & -    & -    & -     & -    & 72.4/72.5 & ~~~~~~-/88.2  & ~~~~~-/\textbf{54.2} & ~~~~~-/63.1 & -    & -    & -    & -   & -   & -   \\
		& VisualBERT \cite{li2019visualbert}  & -    & -    & -     & -    & 70.8/71.0 & -   & -   & -   & -    & -    & -    & 70.8/71.6 & 73.2/73.2 & 52.2/52.4  \\
		& ViLBERT \cite{lu2019vilbert}     & -    & -    & -     & -    & 70.6/70.9 & -   & -   & -   & 58.2 & 84.9 & 91.5 & 72.4/73.3 & 74.5/74.6 & 54.0/54.8  \\
        & VL-BERT \cite{su2019vl}     & -    & -    & -     & -    & 71.2/72.2 & -   & -   & -   & -    & -    & -    & 73.8/-~~~~~~ & 74.4/-~~~~~~ & 55.2/-~~~~~~  \\ \hline
\multirow{4}{*}{\textbf{Pre-ED}}
        & Unified VLP \cite{zhou2019unified} $^\star$ & 39.5 & 29.3 & 129.3 & 23.2 & 70.5/70.7 & 87.2/87.4 & 52.1/52.1 & 60.3/60.5 & -    & -    & -    & -   & -   & -    \\ \cline{2-16}
        & TDEN (Single-stream) & 38.6 & 28.8 & 128.6 & 22.6 & 70.7/-~~~~~~ & 86.9/-~~~~~~ & 53.7/-~~~~~~ & 60.6/-~~~~~~ & 56.2 & 84.3 & 90.3 & 73.3/-~~~~~~ & 74.9/-~~~~~~ & 55.2/-~~~~~~  \\
        & TDEN (Two-stream) & 39.1 & 29.1 & 129.6 & 22.9 & 70.9/-~~~~~~ & 87.2/-~~~~~~ & 52.8/-~~~~~~ & 61.0/-~~~~~~ & 59.2 & 85.4 & 90.9 & 73.9/-~~~~~~ & 75.3/-~~~~~~ & 55.8/-~~~~~~  \\
        & TDEN (Two-stream decoupled)    & \textbf{40.2}/70.7 & \textbf{29.7}/38.7 & \textbf{133.4}/130.4 & \textbf{23.5} & \textbf{72.5}/\textbf{72.8} & \textbf{88.5}/\textbf{88.8} & \textbf{54.7}/53.8 & \textbf{63.0}/\textbf{63.2} & \textbf{63.6} & \textbf{88.2} & \textbf{92.9} & \textbf{75.2}/\textbf{75.7} & \textbf{76.7}/\textbf{76.4} & \textbf{58.1}/\textbf{58.0} \\
  \Xhline{2\arrayrulewidth}
\end{tabular}
\end{table*}

\noindent\textbf{Finetuning Data and Details on Downstream Tasks.}
In \textbf{Visual Question Answering (VQA)},~the model predicts an answer to the given question with regard to an image.
VQA 2.0~\cite{antol2015vqa}~is adopted for finetuning our TDEN, which consists of 1.1 million questions about images in~COCO~\cite{chen2015microsoft}.
During finetuning, we follow the official split \cite{anderson2017bottom} and formulate this task as a multi-label classification problem.
In particular, by feeding the input image-question pair into TDEN, we learn the holistic image-question feature based on the multi-modal outputs of cross-modal encoder \& decoder via attention mechanism \cite{zhouyucvpr2019}. The holistic image-question feature is further embedded into 3,129 possible answers via a fully-connected layer with sigmoid function. As in \cite{anderson2017bottom}, we optimize the output answer predictions with regard to the soft answer labels via cross-entropy loss (mini-batch size: 96, learning rate: 0.00005, maximum iteration: 20 epoches).

\textbf{Caption-based image retrieval (CIR)} aims to search an image from a pool given its caption that depicts the image content.
We utilize Flickr30k \cite{plummer2015flickr30k} in this task and each image is equipped with five human-annotated sentences.
We follow the commonly adopted split in \cite{lee2018stacked} and formulate this task as a ranking problem that sorts images according to the image-sentence similarities, which are measured as in ISM.
The whole model is optimized with triplet ranking loss (mini-batch size: 512, learning rate: 0.00002, maximum iteration: 30 epoches).

\textbf{Visual commonsense reasoning (VCR)} tackles two problems: visual question answering (Q$\to$A) and answer justification (QA$\to$R), that requires the model to predict an answer or judge the correctness of the chosen rationale respectively. Each problem is framed as multiple choice task. In addition, VCR includes a holistic setting (Q$\to$AR) that the model should choose the right answer (from four answer choices) and then select the correct rationale for that answer (from four rationale choices).
The Visual Commonsense Reasoning (VCR) benchmark \cite{zellers2019recognition} is utilized for evaluation.
During finetuning, we concatenate the question and each possible response (answer/rationale) as the input sentence, which is fed into TDEN along with the image.
As in VQA, we obtain the holistic image-sentence feature and then utilize a linear layer to predict the score for each possible response.
The final prediction (i.e., all scores of the four response choices) is thus trained with cross-entropy loss (mini-batch size: 64, learning rate: 0.00002, maximum iteration: 20 epoches).

\textbf{Image captioning (IC)} aims to generate the natural sentence that depicts input image.
COCO \cite{chen2015microsoft} is utilized for fine-tuning and evaluating TDEN.
We utilize the widely adopted Karpathy split \cite{Karpathy:CVPR15,yao2017boosting,yao2018exploring,yao2019hierarchy} for evaluation.
For finetuning, we firstly optimize the whole architecture with cross-entropy loss. The mini-batch size is 16 and the learning rate is 0.00003. We set the maximum iteration as 10 epoches.
The fine-tuned TDEN is further trained with self-critical training strategy \cite{rennie2017self}, which enables the sequence-level optimization with CIDEr reward. The learning rate is 0.000005 and the maximum iteration is 30~epoches.

\begin{table*}[!tb]\small
\centering
\setlength{\extrarowheight}{0.0pt}
\setlength\tabcolsep{1.2pt}
\caption{\small Ablation study on the use of different VL proxy tasks and scheduled sampling for VL pretraining. * indicates a different implementation of ISM based on the outputs of cross-modal encoder.}
\label{table:ablation}
\begin{tabular}{c|ccccc|cccc|cccc|ccc|ccc}
\Xhline{2\arrayrulewidth}
\multirow{2}{*}{\textbf{\#}}&\multirow{2}{*}{\textbf{MLM}} & \multirow{2}{*}{\textbf{MOC}} &\multirow{2}{*}{\textbf{MSG}} & \multirow{2}{*}{\textbf{ISM}} & \multirow{2}{*}{\textbf{Scheduled Sampling}} & \multicolumn{4}{c|}{Image Captioning} & \multicolumn{4}{c|}{VQA (test-dev)} & \multicolumn{3}{c|}{CIR} & \multicolumn{3}{c}{VCR} \\
&&& &  &  & B@4 & M & C & S & Overall & Yes/No & Number  & Other & R1 & R5 & R10 & Q$\rightarrow$A & QA$\rightarrow$R & Q$\rightarrow$AR \\ \hline\hline
1&$\checkmark$&$\checkmark$ & & $\checkmark$ &  & 39.2 & 28.9 & 129.1 & 22.8 & 71.9 & 88.1 & 53.7 & 62.3 & 61.8 & 87.1 & 92.2 & 74.6 & 75.9 & 56.9  \\
2&& & $\checkmark$ & $\checkmark$ &  & 39.3 & 29.3 & 131.0 & 23.1 & 71.7 & 87.9 & 53.4 & 62.0 & 61.7 & 87.0 & 92.2 & 74.5 & 75.9 & 56.8  \\\hline
3&$\checkmark$&$\checkmark$ & $\checkmark$ &  &  & 39.7 & 29.3 & 131.8 & 23.1 & 72.0 & 88.1 & 54.3 & 62.2 & 60.6 & 86.1 & 91.4 & 74.8 & 76.0 & 57.3  \\
4&$\checkmark$&$\checkmark$ & $\checkmark$ & * &  & 39.5 & 29.2 & 130.3 & 23.0 & 71.9 & 88.0 & 54.2 & 62.2 & 60.2 & 85.7 & 91.2 & 74.6 & 76.4 & 57.2  \\ \hline
5&$\checkmark$&$\checkmark$ & $\checkmark$ & $\checkmark$ &   & 39.7 & 29.4 & 132.4 & 23.2 & 72.1 & 88.1 & 54.2 & 62.6 & 62.5 & 86.7 & 92.3 & 74.8 & 76.1 & 57.3  \\ \hline
6&$\checkmark$&$\checkmark$ & $\checkmark$ & $\checkmark$ & Two-Pass-A & 40.2 & 29.7 & 133.4 & 23.5 & 72.5 & 88.5 & 54.7 & 63.0 & 63.6 & 88.2 & 92.9 & 75.2 & 76.7 & 58.1  \\
7&$\checkmark$&$\checkmark$ & $\checkmark$ & $\checkmark$ & Two-Pass-B & 40.1 & 29.6 & 133.0 & 23.4 & 72.3 & 88.4 & 54.2 & 62.7 & 63.6 & 87.8 & 92.9 & 75.3 & 76.9 & 58.2  \\
8&$\checkmark$&$\checkmark$ & $\checkmark$ & $\checkmark$ & Two-Pass-C & 40.0 & 29.6 & 133.0 & 23.4 & 72.2 & 88.1 & 54.6 & 62.7 & 63.4 & 87.4 & 92.7 & 75.1 & 77.0 & 58.3  \\

  \Xhline{2\arrayrulewidth}
  \vspace{-0.3in}
\end{tabular}
\end{table*}

\subsection{Performance Comparison}
Table \ref{table:exp} summarizes the quantitative results of our TDEN on four VL understanding (VQA, CIR, and VCR) and generation (IC) downstream tasks.
We compare TDEN with state-of-the-art task-specific models and VL pre-training approaches (including both pre-trainable encoder and pre-trainable encoder-decoder models) in each downstream task.
Note that we additionally include three variants of our TDEN (with two-stream decoupled structure): (i) \textbf{TDEN$^-$} is directly trained with task-specific training data for each task, without pretraining on Conceptual Captions; (ii) \textbf{TDEN (Single-stream)} is implemented as a shared cross-modal encoder-decoder structure across different modalities and proxy tasks; (iii) \textbf{TDEN (Two-stream)} consists of two object and sentence encoders, followed by a shared cross-modal encoder-decoder across different proxy tasks.

\textbf{Comparison with SOTA Task-specific Models.}
In general, under the same task-specific training setting without VL pre-training, our TDEN$^-$ achieves comparable results with other SOTA baselines on all downstream tasks.
The results basically demonstrate the effectiveness of the adopted two-stream decoupled Transformer based encoder-decoder structure.
Furthermore, when pretraining TDEN on Conceptual Captions and then finetuning it on task-specific data, our TDEN consistently exhibits better performances than other SOTA task-specific baselines across the most evaluation metrics on four tasks.
The performance improvements generally demonstrate the key advantage of exploiting VL pretraining via TDEN, that facilitates both VL understanding and generation downstream tasks.

\textbf{Comparison with VL Pretraining Approaches.}
Overall, the results across all VL understanding and generation tasks consistently indicate that our TDEN exhibits better performances that other state-of-the-art pre-trainable encoder modules (e.g., ViLBERT and LXMERT) and encoder-decoder structure (Unified VLP).
In the IC and VQA tasks, the CIDEr and Overall of TDEN can achieve 133.4\% and 72.5\%, making 4.1\% and 1.3\% absolute improvements over the best competitors Unified VLP and VL-BERT, respectively.
In particular, all the pre-trainable encoder modules lead to a large performance boost over SOTA task-specific approaches for VL understanding tasks. However, the pre-trainable encoder modules can not be directly adapted to VL generation task (IC), which needs an additional language decoder for sentence generation.
By enabling the simultaneous pretraining of encoder and decoder, Unified VLP outperforms SOTA task-specific approaches on VL generation task. However, the results of Unified VLP on VL understanding tasks are still lower than those of pre-trainable encoders (e.g., VL-BERT).
The same observation is also noticed when comparing our TDEN (Single-stream) and TDEN (Two-stream) against VL-BERT. We speculate that the performance degradations on VL understanding tasks may be caused by the joint pre-training with two different kinds of VL proxy tasks (MLM and MSG) in a shared encoder-decoder structure.
Furthermore, by decoupling the two kinds of proxy tasks with two decoupled cross-modal encoder and decoder during pretraining, our TDEN boosts the performances in all the VL understanding and generation downstream tasks. This confirms the effectiveness of two-stream decoupled design in our TDEN.

\subsection{Ablation Analysis}
To better understand our TDEN, we perform an ablation analysis in Table \ref{table:ablation}.

\textbf{Pretraining Cross-modal Encoder and Decoder Simultaneously v.s. Separately.}
We tried pretraining the cross-modal encoder and decoder separately. In particular, the cross-modal encoder is pretrained with ISM, MLM, and MOC, which is only applied for VL understanding downstream tasks. Meanwhile, we pretrain the cross-modal decoder with ISM and MSG, which is specifically utilized for VL generation downstream task. When separating pretraining process, the final results on four downstream tasks (Row 1\&2 in Table \ref{table:ablation}) are lower than simultaneous pretraining (Row 5 in Table \ref{table:ablation}).
This verifies the merit of simultaneous pretraining strategy, which jointly exploits the underlying common vision-language associations across understanding and generation proxy tasks.

\textbf{Performing ISM with Object/Sentence Encoders v.s. Cross-modal Encoder.}
We also experimented by using the multi-modal outputs of cross-modal encoder for ISM as in \cite{lu2019vilbert}. The performances on all tasks (Row 4 in Table \ref{table:ablation}) are even lower than our design without ISM (Row 3 in Table \ref{table:ablation}). The results align with the observation in \cite{su2019vl} and show that the typical design of ISM hampers the pretrianing of other proxy tasks by introducing mismatched image-sentence pairs in the shared cross-modal encoder. Instead, by capitalizing on the outputs of object and sentence encoders for ISM, our design (Row 5 in Table \ref{table:ablation}) boosts up the performances. This basically validates the effectiveness of an earlier image-sentence alignment via our adopted ISM.

\textbf{Comparison of Different Two-pass Schemes for Scheduled Sampling.}
We finally examine how performances on downstream tasks are affected when capitalizing on different two-pass schemes for scheduled sampling in pretraining.
Overall, in our TDEN, all the three two-pass schemes (Row 6-8 in Table \ref{table:ablation}) exhibit better performance than the one without any scheduled sampling strategy (Row 5 in Table \ref{table:ablation}), which clearly indicates the advantage of exploring scheduled sampling for VL pretraining. Moreover, compared to Two-Pass-A and Two-Pass-B, Two-Pass-C effectively reduces the pretraining cost with only a slight performance drop across all tasks, which is a very practical choice.

\section{Conclusions}
In this paper, we developed TDEN, a new pre-trainable encoder-decoder structure that simultaneously supports both vision-language understanding and generation downstream tasks. Particularly, instead of using the common single-stream or two-stream architecture for pretraining, TDEN capitalizes on a two-stream decoupled design of encoder-decoder to exploit the mutual but also fuzzy relations across different modalities and proxy tasks. Furthermore, a primary scheduled sampling strategy is novelly proposed to elegantly mitigate the pretraining-finetuning discrepancy for our TDEN pretraining.
Extensive experiments demonstrate the compelling generalizability of TDEN by fine-tuning it to four vision-language understanding and generation tasks. More remarkably, TDEN pretraining leads to a new state-of-the-art performance on each downstream task.

\section{Ethical Statement}
Vision-language technologies (e.g., visual question answering and image captioning) have a great potential impact for instance on robotic vision or helping visually impaired people. Nevertheless, the achievements of these technologies rely heavily on the requirement of large quantities of task-specific annotations (e.g., image-question-answer triplets/image-sentence pairs) for neural model learning. This severely hinders the scalability and generalization of vision-language techniques when only limited annotations are available. To alleviate this problem, we focus on pretraining a multi-modal encoder-decoder model over an automatically collected large-scale image captioning dataset without any human annotations, which could facilitate a wide range of vision-language applications (e.g., visual question answering, caption-based image retrieval, visual commonsense reasoning, and image captioning).
However, one potential risk lies in that if the use of vision-language pretraining means vision-language understanding/generation systems may now be easily developed by those with lower levels of domain or ML expertise, this could increase the risk of the vision-language model or its outputs being used incorrectly.

\bibliographystyle{aaai21}
\bibliography{egbib}

\begin{thebibliography}{48}
\providecommand{\natexlab}[1]{#1}
\providecommand{\url}[1]{\texttt{#1}}
\providecommand{\urlprefix}{URL }
\expandafter\ifx\csname urlstyle\endcsname\relax
  \providecommand{\doi}[1]{doi:\discretionary{}{}{}#1}\else
  \providecommand{\doi}{doi:\discretionary{}{}{}\begingroup
  \urlstyle{rm}\Url}\fi

\bibitem[{Anderson et~al.(2018)Anderson, He, Buehler, Teney, Johnson, Gould,
  and Zhang}]{anderson2017bottom}
Anderson, P.; He, X.; Buehler, C.; Teney, D.; Johnson, M.; Gould, S.; and
  Zhang, L. 2018.
\newblock Bottom-up and top-down attention for image captioning and visual
  question answering.
\newblock In \emph{CVPR}.

\bibitem[{Antol et~al.(2015)Antol, Agrawal, Lu, Mitchell, Batra,
  Lawrence~Zitnick, and Parikh}]{antol2015vqa}
Antol, S.; Agrawal, A.; Lu, J.; Mitchell, M.; Batra, D.; Lawrence~Zitnick, C.;
  and Parikh, D. 2015.
\newblock Vqa: Visual question answering.
\newblock In \emph{ICCV}.

\bibitem[{Bengio et~al.(2015)Bengio, Vinyals, Jaitly, and
  Shazeer}]{bengio2015scheduled}
Bengio, S.; Vinyals, O.; Jaitly, N.; and Shazeer, N. 2015.
\newblock Scheduled sampling for sequence prediction with recurrent neural
  networks.
\newblock In \emph{NeurIPS}.

\bibitem[{Chen et~al.(2019{\natexlab{a}})Chen, Pan, Li, Yao, Chao, and
  Mei}]{chen2019temporal}
Chen, J.; Pan, Y.; Li, Y.; Yao, T.; Chao, H.; and Mei, T. 2019{\natexlab{a}}.
\newblock Temporal deformable convolutional encoder-decoder networks for video
  captioning.
\newblock In \emph{AAAI}.

\bibitem[{Chen et~al.(2015)Chen, Fang, Lin, Vedantam, Gupta, Doll{\'a}r, and
  Zitnick}]{chen2015microsoft}
Chen, X.; Fang, H.; Lin, T.-Y.; Vedantam, R.; Gupta, S.; Doll{\'a}r, P.; and
  Zitnick, C.~L. 2015.
\newblock Microsoft {COCO} captions: Data collection and evaluation server.
\newblock \emph{arXiv preprint arXiv:1504.00325} .

\bibitem[{Chen et~al.(2019{\natexlab{b}})Chen, Li, Yu, Kholy, Ahmed, Gan,
  Cheng, and Liu}]{chen2019uniter}
Chen, Y.-C.; Li, L.; Yu, L.; Kholy, A.~E.; Ahmed, F.; Gan, Z.; Cheng, Y.; and
  Liu, J. 2019{\natexlab{b}}.
\newblock Uniter: Learning universal image-text representations.
\newblock \emph{arXiv preprint arXiv:1909.11740} .

\bibitem[{Deng et~al.(2009)Deng, Dong, Socher, Li, Li, and Fei-Fei.}]{ImageNet}
Deng, J.; Dong, W.; Socher, R.; Li, L.-J.; Li, K.; and Fei-Fei., L. 2009.
\newblock ImageNet: A Large-Scale Hierarchical Image Database.
\newblock In \emph{CVPR}.

\bibitem[{Devlin et~al.(2019)Devlin, Chang, Lee, and
  Toutanova}]{devlin2019bert}
Devlin, J.; Chang, M.-W.; Lee, K.; and Toutanova, K. 2019.
\newblock BERT: Pre-training of Deep Bidirectional Transformers for Language
  Understanding.
\newblock In \emph{NAACL}.

\bibitem[{Dong et~al.(2019)Dong, Yang, Wang, Wei, Liu, Wang, Gao, Zhou, and
  Hon}]{dong2019unified}
Dong, L.; Yang, N.; Wang, W.; Wei, F.; Liu, X.; Wang, Y.; Gao, J.; Zhou, M.;
  and Hon, H.-W. 2019.
\newblock Unified language model pre-training for natural language
  understanding and generation.
\newblock In \emph{NeurIPS}.

\bibitem[{Gao et~al.(2019)Gao, Jiang, You, Lu, Hoi, Wang, and
  Li}]{gao2019dynamic}
Gao, P.; Jiang, Z.; You, H.; Lu, P.; Hoi, S.~C.; Wang, X.; and Li, H. 2019.
\newblock Dynamic fusion with intra-and inter-modality attention flow for
  visual question answering.
\newblock In \emph{CVPR}.

\bibitem[{Huang et~al.(2019)Huang, Wang, Chen, and Wei}]{huang2019attention}
Huang, L.; Wang, W.; Chen, J.; and Wei, X.-Y. 2019.
\newblock Attention on attention for image captioning.
\newblock In \emph{ICCV}.

\bibitem[{Karpathy and Fei-Fei(2015)}]{Karpathy:CVPR15}
Karpathy, A.; and Fei-Fei, L. 2015.
\newblock Deep Visual-Semantic Alignments for Generating Image Descriptions.
\newblock In \emph{CVPR}.

\bibitem[{Kim, Jun, and Zhang(2018)}]{kim2018bilinear}
Kim, J.-H.; Jun, J.; and Zhang, B.-T. 2018.
\newblock Bilinear attention networks.
\newblock In \emph{NeurIPS}.

\bibitem[{Kingma and Ba(2015)}]{kingma2014adam}
Kingma, D.; and Ba, J. 2015.
\newblock Adam: A method for stochastic optimization.
\newblock In \emph{ICLR}.

\bibitem[{Krishna et~al.(2017)Krishna, Zhu, Groth, Johnson, Hata, Kravitz,
  Chen, Kalantidis, Li, Shamma et~al.}]{krishna2017visual}
Krishna, R.; Zhu, Y.; Groth, O.; Johnson, J.; Hata, K.; Kravitz, J.; Chen, S.;
  Kalantidis, Y.; Li, L.-J.; Shamma, D.~A.; et~al. 2017.
\newblock Visual genome: Connecting language and vision using crowdsourced
  dense image annotations.
\newblock \emph{IJCV} .

\bibitem[{Lee et~al.(2018)Lee, Chen, Hua, Hu, and He}]{lee2018stacked}
Lee, K.-H.; Chen, X.; Hua, G.; Hu, H.; and He, X. 2018.
\newblock Stacked cross attention for image-text matching.
\newblock In \emph{ECCV}.

\bibitem[{Lewis et~al.(2019)Lewis, Liu, Goyal, Ghazvininejad, Mohamed, Levy,
  Stoyanov, and Zettlemoyer}]{lewis2019bart}
Lewis, M.; Liu, Y.; Goyal, N.; Ghazvininejad, M.; Mohamed, A.; Levy, O.;
  Stoyanov, V.; and Zettlemoyer, L. 2019.
\newblock Bart: Denoising sequence-to-sequence pre-training for natural
  language generation, translation, and comprehension.
\newblock \emph{arXiv preprint arXiv:1910.13461} .

\bibitem[{Li et~al.(2019{\natexlab{a}})Li, Duan, Fang, Jiang, and
  Zhou}]{li2019unicoder}
Li, G.; Duan, N.; Fang, Y.; Jiang, D.; and Zhou, M. 2019{\natexlab{a}}.
\newblock Unicoder-vl: A universal encoder for vision and language by
  cross-modal pre-training.
\newblock \emph{arXiv preprint arXiv:1908.06066} .

\bibitem[{Li et~al.(2019{\natexlab{b}})Li, Yatskar, Yin, Hsieh, and
  Chang}]{li2019visualbert}
Li, L.~H.; Yatskar, M.; Yin, D.; Hsieh, C.-J.; and Chang, K.-W.
  2019{\natexlab{b}}.
\newblock Visualbert: A simple and performant baseline for vision and language.
\newblock \emph{arXiv preprint arXiv:1908.03557} .

\bibitem[{Li et~al.(2018)Li, Yao, Pan, Chao, and Mei}]{li2018jointly}
Li, Y.; Yao, T.; Pan, Y.; Chao, H.; and Mei, T. 2018.
\newblock Jointly localizing and describing events for dense video captioning.
\newblock In \emph{CVPR}.

\bibitem[{Li et~al.(2019{\natexlab{c}})Li, Yao, Pan, Chao, and
  Mei}]{li2019pointing}
Li, Y.; Yao, T.; Pan, Y.; Chao, H.; and Mei, T. 2019{\natexlab{c}}.
\newblock Pointing novel objects in image captioning.
\newblock In \emph{CVPR}.

\bibitem[{Lu et~al.(2019)Lu, Batra, Parikh, and Lee}]{lu2019vilbert}
Lu, J.; Batra, D.; Parikh, D.; and Lee, S. 2019.
\newblock Vilbert: Pretraining task-agnostic visiolinguistic representations
  for vision-and-language tasks.
\newblock In \emph{NeurIPS}.

\bibitem[{Pan et~al.(2020{\natexlab{a}})Pan, Li, Luo, Xu, Yao, and
  Mei}]{pan2020auto}
Pan, Y.; Li, Y.; Luo, J.; Xu, J.; Yao, T.; and Mei, T. 2020{\natexlab{a}}.
\newblock Auto-captions on GIF: A Large-scale Video-sentence Dataset for
  Vision-language Pre-training.
\newblock \emph{arXiv preprint arXiv:2007.02375} .

\bibitem[{Pan et~al.(2016)Pan, Mei, Yao, Li, and Rui}]{pan2016jointly}
Pan, Y.; Mei, T.; Yao, T.; Li, H.; and Rui, Y. 2016.
\newblock Jointly modeling embedding and translation to bridge video and
  language.
\newblock In \emph{CVPR}.

\bibitem[{Pan et~al.(2017)Pan, Yao, Li, and Mei}]{pan2017video}
Pan, Y.; Yao, T.; Li, H.; and Mei, T. 2017.
\newblock Video captioning with transferred semantic attributes.
\newblock In \emph{CVPR}.

\bibitem[{Pan et~al.(2020{\natexlab{b}})Pan, Yao, Li, and Mei}]{pan2020x}
Pan, Y.; Yao, T.; Li, Y.; and Mei, T. 2020{\natexlab{b}}.
\newblock X-Linear Attention Networks for Image Captioning.
\newblock In \emph{CVPR}.

\bibitem[{Paszke et~al.(2019)Paszke, Gross, Massa, Lerer, Bradbury, Chanan,
  Killeen, Lin, Gimelshein, Antiga et~al.}]{paszke2019pytorch}
Paszke, A.; Gross, S.; Massa, F.; Lerer, A.; Bradbury, J.; Chanan, G.; Killeen,
  T.; Lin, Z.; Gimelshein, N.; Antiga, L.; et~al. 2019.
\newblock PyTorch: An imperative style, high-performance deep learning library.
\newblock In \emph{NeurIPS}.

\bibitem[{Peters et~al.(2018)Peters, Neumann, Iyyer, Gardner, Clark, Lee, and
  Zettlemoyer}]{peters2018deep}
Peters, M.~E.; Neumann, M.; Iyyer, M.; Gardner, M.; Clark, C.; Lee, K.; and
  Zettlemoyer, L. 2018.
\newblock Deep contextualized word representations.
\newblock In \emph{NAACL}.

\bibitem[{Plummer et~al.(2015)Plummer, Wang, Cervantes, Caicedo, Hockenmaier,
  and Lazebnik}]{plummer2015flickr30k}
Plummer, B.~A.; Wang, L.; Cervantes, C.~M.; Caicedo, J.~C.; Hockenmaier, J.;
  and Lazebnik, S. 2015.
\newblock Flickr30k entities: Collecting region-to-phrase correspondences for
  richer image-to-sentence models.
\newblock In \emph{ICCV}.

\bibitem[{Radford et~al.(2018)Radford, Narasimhan, Salimans, and
  Sutskever}]{radford2018improving}
Radford, A.; Narasimhan, K.; Salimans, T.; and Sutskever, I. 2018.
\newblock Improving language understanding by generative pre-training.
\newblock \emph{Technical report, OpenAI} .

\bibitem[{Ren et~al.(2015)Ren, He, Girshick, and Sun}]{ren2015faster}
Ren, S.; He, K.; Girshick, R.; and Sun, J. 2015.
\newblock Faster r-cnn: Towards real-time object detection with region proposal
  networks.
\newblock In \emph{NeurIPS}.

\bibitem[{Rennie et~al.(2017)Rennie, Marcheret, Mroueh, Ross, and
  Goel}]{rennie2017self}
Rennie, S.~J.; Marcheret, E.; Mroueh, Y.; Ross, J.; and Goel, V. 2017.
\newblock Self-Critical Sequence Training for Image Captioning.
\newblock In \emph{CVPR}.

\bibitem[{Sharma et~al.(2018)Sharma, Ding, Goodman, and
  Soricut}]{sharma2018conceptual}
Sharma, P.; Ding, N.; Goodman, S.; and Soricut, R. 2018.
\newblock Conceptual captions: A cleaned, hypernymed, image alt-text dataset
  for automatic image captioning.
\newblock In \emph{ACL}.

\bibitem[{Song et~al.(2019)Song, Tan, Qin, Lu, and Liu}]{song2019mass}
Song, K.; Tan, X.; Qin, T.; Lu, J.; and Liu, T.-Y. 2019.
\newblock MASS: Masked Sequence to Sequence Pre-training for Language
  Generation.
\newblock In \emph{ICML}.

\bibitem[{Su et~al.(2019)Su, Zhu, Cao, Li, Lu, Wei, and Dai}]{su2019vl}
Su, W.; Zhu, X.; Cao, Y.; Li, B.; Lu, L.; Wei, F.; and Dai, J. 2019.
\newblock Vl-bert: Pre-training of generic visual-linguistic representations.
\newblock \emph{arXiv preprint arXiv:1908.08530} .

\bibitem[{Tadmor et~al.(2016)Tadmor, Rosenwein, Shalev-Shwartz, Wexler, and
  Shashua}]{tadmor2016learning}
Tadmor, O.; Rosenwein, T.; Shalev-Shwartz, S.; Wexler, Y.; and Shashua, A.
  2016.
\newblock Learning a Metric Embedding for Face Recognition using the Multibatch
  Method.
\newblock In \emph{NeurIPS}.

\bibitem[{Tan and Bansal(2019)}]{tan2019lxmert}
Tan, H.; and Bansal, M. 2019.
\newblock LXMERT: Learning Cross-Modality Encoder Representations from
  Transformers.
\newblock In \emph{EMNLP-IJCNLP}.

\bibitem[{Vaswani et~al.(2017)Vaswani, Shazeer, Parmar, Uszkoreit, Jones,
  Gomez, Kaiser, and Polosukhin}]{vaswani2017attention}
Vaswani, A.; Shazeer, N.; Parmar, N.; Uszkoreit, J.; Jones, L.; Gomez, A.~N.;
  Kaiser, {\L}.; and Polosukhin, I. 2017.
\newblock Attention is all you need.
\newblock In \emph{NeurIPS}.

\bibitem[{Vinyals et~al.(2015)Vinyals, Toshev, Bengio, and
  Erhan}]{vinyals2015show}
Vinyals, O.; Toshev, A.; Bengio, S.; and Erhan, D. 2015.
\newblock Show and tell: A neural image caption generator.
\newblock In \emph{CVPR}.

\bibitem[{Xu et~al.(2016)Xu, Mei, Yao, and Rui}]{xu2016msr}
Xu, J.; Mei, T.; Yao, T.; and Rui, Y. 2016.
\newblock Msr-vtt: A large video description dataset for bridging video and
  language.
\newblock In \emph{CVPR}.

\bibitem[{Yang et~al.(2019)Yang, Dai, Yang, Carbonell, Salakhutdinov, and
  Le}]{yang2019xlnet}
Yang, Z.; Dai, Z.; Yang, Y.; Carbonell, J.; Salakhutdinov, R.~R.; and Le, Q.~V.
  2019.
\newblock XLNet: Generalized Autoregressive Pretraining for Language
  Understanding.
\newblock In \emph{NeurIPS}.

\bibitem[{Yao et~al.(2017{\natexlab{a}})Yao, Pan, Li, and
  Mei}]{yao2017incorporating}
Yao, T.; Pan, Y.; Li, Y.; and Mei, T. 2017{\natexlab{a}}.
\newblock Incorporating copying mechanism in image captioning for learning
  novel objects.
\newblock In \emph{CVPR}.

\bibitem[{Yao et~al.(2018)Yao, Pan, Li, and Mei}]{yao2018exploring}
Yao, T.; Pan, Y.; Li, Y.; and Mei, T. 2018.
\newblock Exploring visual relationship for image captioning.
\newblock In \emph{ECCV}.

\bibitem[{Yao et~al.(2019)Yao, Pan, Li, and Mei}]{yao2019hierarchy}
Yao, T.; Pan, Y.; Li, Y.; and Mei, T. 2019.
\newblock Hierarchy parsing for image captioning.
\newblock In \emph{ICCV}.

\bibitem[{Yao et~al.(2017{\natexlab{b}})Yao, Pan, Li, Qiu, and
  Mei}]{yao2017boosting}
Yao, T.; Pan, Y.; Li, Y.; Qiu, Z.; and Mei, T. 2017{\natexlab{b}}.
\newblock Boosting image captioning with attributes.
\newblock In \emph{ICCV}.

\bibitem[{Yu et~al.(2019)Yu, Yu, Cui, Tao, and Tian}]{zhouyucvpr2019}
Yu, Z.; Yu, J.; Cui, Y.; Tao, D.; and Tian, Q. 2019.
\newblock Deep modular co-attention networks for visual question answering.
\newblock In \emph{CVPR}.

\bibitem[{Zellers et~al.(2019)Zellers, Bisk, Farhadi, and
  Choi}]{zellers2019recognition}
Zellers, R.; Bisk, Y.; Farhadi, A.; and Choi, Y. 2019.
\newblock From recognition to cognition: Visual commonsense reasoning.
\newblock In \emph{CVPR}.

\bibitem[{Zhou et~al.(2020)Zhou, Palangi, Zhang, Hu, Corso, and
  Gao}]{zhou2019unified}
Zhou, L.; Palangi, H.; Zhang, L.; Hu, H.; Corso, J.~J.; and Gao, J. 2020.
\newblock Unified vision-language pre-training for image captioning and vqa.
\newblock In \emph{AAAI}.

\end{thebibliography}

\end{document}